\documentclass{article}
\usepackage{DFAM_DETR}

\usepackage{times}
\usepackage{soul}
\usepackage{url}
\usepackage[utf8]{inputenc}
\usepackage[small]{caption}
\usepackage{graphicx}
\usepackage{amsmath}
\usepackage{amsthm}
\usepackage{booktabs}
\usepackage{algorithm}
\usepackage{algorithmic}
\usepackage{amssymb}

\title{DFAM-DETR: Deformable feature based attention mechanism DETR on slender object detection}

\author{
Wen Feng$^1$,
Wang Mei$^1$,
Hu Xiaojie$^{1}$\footnote{Contact Author}
\affiliations
$^1$Shenyang Ligong University\\
\emails
\ xiaojie.hu@sylu.edu.cn
}

\begin{document}

\maketitle

\begin{abstract}
Object detection is one of the most significant aspects of computer vision, and it has achieved substantial results in a variety of domains. It is worth noting that there are few studies focusing on slender object detection. CNNs are widely employed in object detection, however it performs poorly on slender object detection due to the fixed geometric structure and sampling points. In comparison, Deformable DETR has the ability to obtain global to specific features. Even though it outperforms the CNNs in slender objects detection accuracy and efficiency, the results are still not satisfactory. Therefore, we propose Deformable Feature based Attention Mechanism (DFAM) to increase the slender object detection accuracy and efficiency of Deformable DETR. The DFAM has adaptive sampling points of deformable convolution and attention mechanism that aggregate information from the entire input sequence in the backbone network. This improved detector is named as Deformable Feature based Attention Mechanism DETR (DFAM-DETR). Results indicate that DFAM-DETR achieves outstanding detection performance on slender objects.
\end{abstract}

\section{Introduction}

\par Object detection has made significant development in recent years as a crucial study subject, with the increasing application of deep learning on computer vision \cite{he2016deep,jiao2019sdlod,zhao2020rdlod,dai2016r}. For each object of interest in an image, object detection needs the algorithm to predict a bounding box with a category label. One-stage detectors and two-stage detectors are the two primary types of object detectors. For instance, the YOLO series \cite{bochkovskiy2020yolov4,redmon2017yolo9000,redmon2018yolov3,redmon2016you}, SSD \cite{liu2016ssd}, DSSD \cite{fu2017dssd}, Retina-Net \cite{lin2017focal}, Efficient-Det \cite{tan2020efficientdet}, FCOS \cite{tian2019fcos}, and Corner-Net \cite{law2018cornernet} are one-stage detectors with the benefit of detection speed. The R-CNN \cite{girshick2014rich}, Cascade R-CNN \cite{cai2018cascade}, Fast R-CNN \cite{girshick2015fast} , and Faster R-CNN \cite{ren2015faster} are two-stage detectors with the advantage of high detection accuracy.
\par The demand for object detection, such as small object detection \cite{zoph2020learning} and dense object detection \cite{huang1509unifying,florence2018dense}, is growing as the field of computer vision develops. Certain detecting effects have been obtained, and some novel approaches and solutions have been offered. Despite the fact that the most of the issues have been resolved, there is still a significant difference between slender object detection and regular object detection. For slender object detection, the previously described one-stage and two-stage detectors, such as Faster R-CNN \cite{ren2015faster}, RepPoint \cite{yang2019reppoints}, and FCOS \cite{tian2019fcos}, were used. The test included only slender object images from the MS-COCO dataset \cite{lin2014microsoft}, such as knives, forks, skis and snowboards. The highest detection accuracy of AP reached 20.7 percent. This is because of the standard convolution only samples the input feature map at fixed locations and it cannot automatically adjust the sampling points to fit the features of slender objects. For instance, as shown in Figure 1(b) the sampling points on slender object is more accurate than the standard convolution in Figure 1(a).

\begin{figure}
	\centering
		\includegraphics[scale=.30]{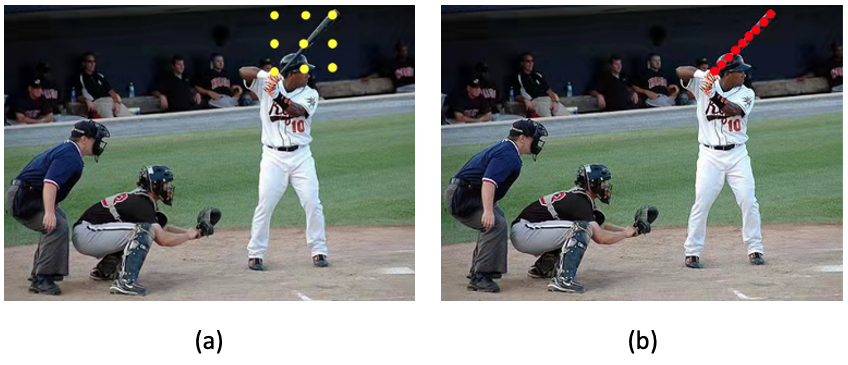}
	\caption{Sampling points for slender objects. a. Standard convolution sampling points; b. Ideal sampling points}
	\label{fig:figure1}
\end{figure}

\par Since Transformer's self-attention layers are global instead of locality two-dimensional neighborhood structure, it has much less image-specific inductive bias than CNNs \cite{dosovitskiy2020image}. Research such as Detection Transformer (DETR) \cite{carion2020end} started to applying transformer for object detection. Results show that the attention mechanism in the transformer has strong modeling capability for relation. The main target area is obtained by scanning the global image. It effectively concentrates on the image's slender object for improved output quality. Furthermore, the data dimension is reduced which can lower the computational load of high-dimension data input.
\par The downside of DETR is that using the transformer attention mechanism to obtain sampling points is still time demanding. Deformable DETR \cite{zhu2021ddteod} successfully integrates transformer and deformable convolution \cite{dai2017deformable} with sparse spatial sampling positions to solve the problem of slow convergence speed and high complexity of DETR. The deformable attention module in Deformable DETR only obtain key sampling points around a reference. Convergence and feature spatial resolution issues can be reduced by allocating a fixed number of important points to each query. It can provide efficient and better detecting performance with fewer and more precise sampling points on the slender objects.
\par Furthermore, while Deformable DETR merely adds a deformable attention module to the transformer, backbone network feature extraction is still insufficient for detecting slender objects. When CNN is used to extract features in the backbone network, it has difficulty adapting to the shape of slender objects. Hence, we propose Deformable Feature based Attention Mechanism DETR (DFAM-DETR) detector for slender object detection. This detector is based on Deformable DETR, and Deformable Feature based Attention Mechanism (DFAM) is designed to sample slender object features and increase the ability of feature extraction by deformable convolution and attention mechanism. Deformable convolution can adjust the position of sample points in the image adaptively. It assures that the sample points are localized in the image's region of interest to avoid background influence. Apart from the deformable convolution, the attention mechanism is also applied with DFAM for slender object detection. The attention mechanism may learn which information to emphasize or suppress based on the dimensions of the channel and space. Hence, it increases the effectiveness of recognizing slender objects by focusing on important features and suppressing those that aren't. 
\par In summary, the proposed DFAM-DETR detector is modified based on Deformable DETR specifically for slender object detection. The DFAM is designed for capturing the specific features of slender objects. As a result, DFAM-DETR detector greatly improves slender object detection accuracy and efficiency comparing to Deformable DETR.

\section{Related work}
\par Object detection is categorized into one-stage detectors and two-stage detectors \cite{he2016deep,jiao2019sdlod,zhao2020rdlod,dai2016r}. One-stage detectors do not require the region proposal stage and may generate the probability of an object's category and location directly. For instance, the YOLO series \cite{bochkovskiy2020yolov4,redmon2017yolo9000,redmon2018yolov3,redmon2016you}, SSD \cite{liu2016ssd}, DSSD \cite{fu2017dssd}, Retina-Net \cite{lin2017focal}, Efficient-Det \cite{tan2020efficientdet}, FCOS \cite{tian2019fcos}, and Corner-Net \cite{law2018cornernet} are typical one-stage detectors with the benefit of detection speed. The two-stage detectors must first create region proposals, then perform object classification and localization for region proposals. For instance, the R-CNN \cite{girshick2014rich}, Cascade R-CNN \cite{cai2018cascade}, Fast R-CNN \cite{girshick2015fast}, and Faster R-CNN \cite{ren2015faster} detectors have the benefit of high detection accuracy.
\par Despite the fact that object detection using convolution has gained high accuracy, detection performance on slender objects remains poor. The convolutional approach has difficulty capturing features of slender objects. Popular object detectors like Faster R-CNN \cite{ren2015faster}, FCOS \cite{tian2019fcos}, RepPoints \cite{yang2019reppoints} adopt standard convolution. Furthermore, improved detectors \cite{wan2020slender} based on FCOS and RepPoints that specifically developed for slender object detection still shows weak detection accuracy.
\par The self-attention mechanisms of transformer can scan through each element of a sequence and update it by aggregating information from the whole sequence \cite{carion2020end,vaswani2017attention,zhu2021ddteod,han2020survey,zhang2021image,yang2021transformer}. DETR is a transformer-based object detector. It combines the bipartite matching loss and transformers with powerful relationship modeling ability \cite{carion2020end}. However, DETR requires more epochs to achieve convergence comparing to popular detectors \cite{zhu2021ddteod}. Due to the complexity of high-resolution feature map, the performance of DETR in detecting small objects is relatively poor \cite{zhu2021ddteod}. Deformable DETR is an effective and efficient detector for dealing with sparse spatial locations, which compensates the lack of the element relation modeling capability for DETR \cite{dai2017deformable,zhu2021ddteod}. The deformable attention module of Deformable DETR only focuses on a small group of key sampling points around the reference point without considering the spatial size of the feature map \cite{zhu2021ddteod}. By allocating only a few fixed numbers of keywords to each query, the problems of convergence and spatial resolution of elements can be alleviated \cite{zhu2021ddteod}.
\par However, Deformable DETR only introduces deformable attention module into transformer. For slender objects, feature extraction of backbone network still adopts convolution, which does not provide sufficient solution. Hence, we propose DFAM based on deformable convolution feature and attention mechanism for effective slender object detection. Unlike Deformable DETR, DFAM use adaptive sampling points of deformable convolution and attention mechanism to aggregate the whole input sequence information in backbone network to accurately identify slender objects and obtain better detection accuracy for slender objects.

\section{Method}
\subsection{Architecture}
\par DFAM-DETR is based on Deformable DETR, which is comprised of three parts, the ResNet \cite{he2016deep} as backbone, the transformer with encoder-decoder, and the Feed Forward Network (FFN). As shown in Figure 2, our improvement is mainly in backbone network. The Deformable Feature based Attention Mechanism (DFAM) is designed in the backbone network based on ResNet to extract slender object features. Transformer takes full advantage of its powerful modeling capabilities and sampling capability of deformable attention module to improve the accuracy of slender object detection. FFN is used to predict the output categories and position of objects in the picture.   

\begin{figure*}[t]
	\centering
	\includegraphics[width=1\textwidth]{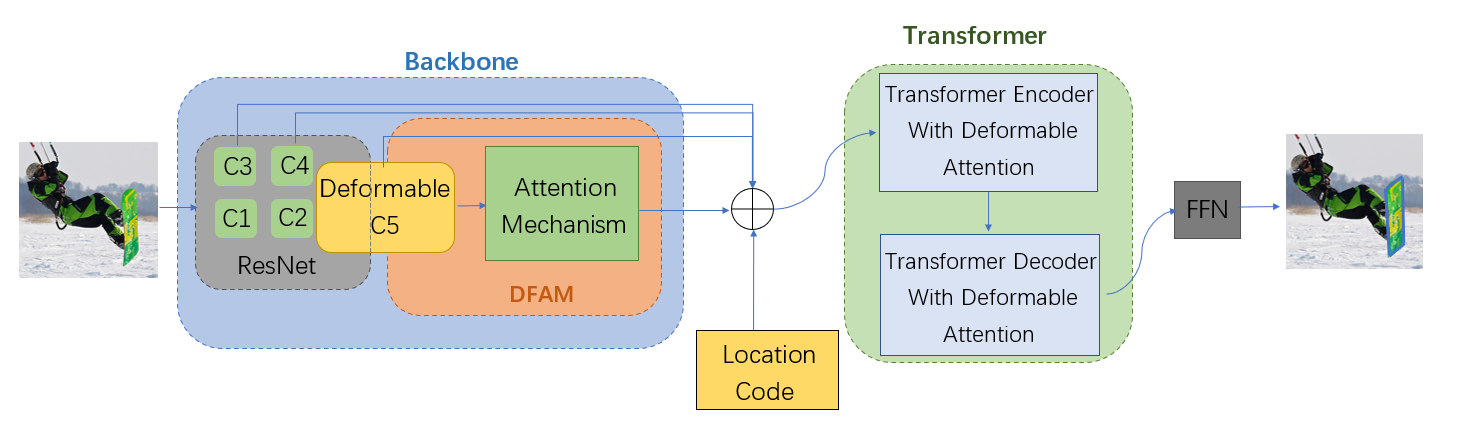}
	\caption{Overall structure of DFAM-DETR. Note: C1, C2, C3 and C4 are original from ResNet.}
	\label{fig:figure2}
\end{figure*}

\subsection{Feature extraction of backbone}
\par The backbone is anticipated to fully mine the meaningful semantic information of the image as the model's core feature extraction function. Convolution in the ResNet backbone is challenging to adapt to the unique shape of slender objects. We propose the DFAM to enhance the ability of feature extraction by improving one layer (C5) in ResNet, see Figure 2. First, ResNet generates feature maps for stages 3, 4, and 5 as C3, C4, C5. To build the deformable C5 feature map, we replace the convolution in the last stage of ResNet with deformable convolution, which is then fed into the attention mechanism to generate the DFAM feature map. Second, in order to capture the different scales of slender objects, the C3, C4, deformable C5 and DFAM feature map are adapted to generate multi-scale slender object feature maps. The C3, C4, and deformable C5 feature maps are convolved with the $ 1\times 1 $ stride 1 to get the first three feature layers. Then the last layer feature map is obtained via a $ 3\times 3 $ stride 2 convolution on the DFAM feature map, see Figure 3. Therefore, the multi-scale slender object feature maps are captured from the backbone. Lastly, the multi-scale slender object feature maps are input to transformer to enhance the ability of semantic and geometric information representation. 

\begin{figure*}[t]
	\centering
	\includegraphics[width=1\textwidth]{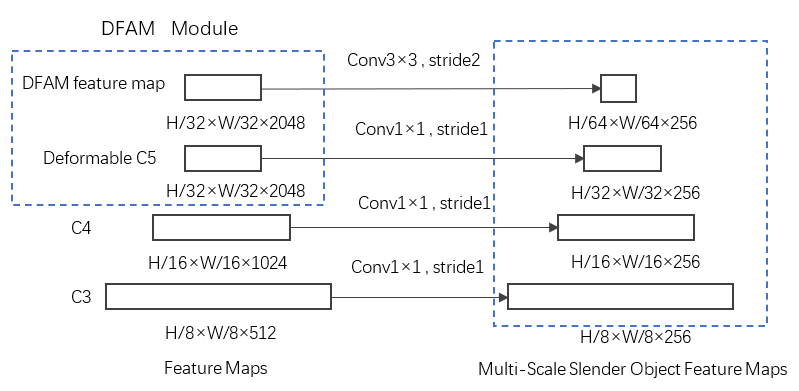}
	\caption{Constructing multi-scale slender object feature maps}
	\label{fig:figure3}
\end{figure*}

\subsection{The proposed Deformable Feature based Attention Mechanism} 
\par As shown in Figure 4, DFAM combines the ability of deformable convolution's adaptive sampling points with the capacity of focusing critical features of the attention mechanism to adjust to the features of slender objects and increase feature extraction ability. 

\begin{figure*}[t]
	\centering
	\includegraphics[width=1\textwidth]{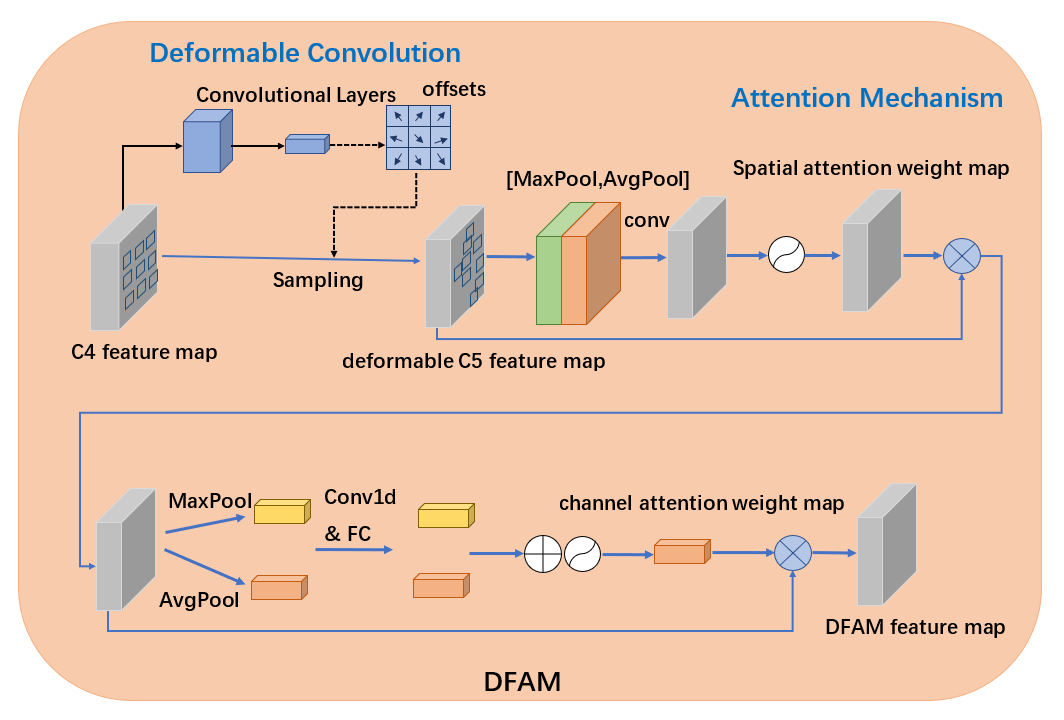}
	\caption{DFAM framework}
	\label{fig:figure4}
\end{figure*}

\par The geometric structure and sample points of the standard convolution kernel are fixed in the convolutional neural network, and the generalization capacity is limited, thus the geometric modification has inherent restrictions. Because of this constraint, a model can only get feature information from a fixed area, making it impossible to adjust to the feature of slender objects and limiting their capacity to extract features. As a result, we propose using deformable convolution instead of standard convolution in the backbone to extract features of slender objects. The comparison between standard convolution sampling points and deformable convolution sampling points in slender object is showing in Figure 5. Comparing to standard convolution, the deformable convolution includes a learnable offset at each sampled position in the feature map so that the deformable convolution can better adapt to the features of slender objects. Furthermore, while deformable convolution does not considerably increase the model's parameters and FLOPS, too many deformable convolution layers would dramatically increase the infer time in reality. As a result, in order to balance efficiency and accuracy, we propose replacing  $ 3\times 3 $ convolution layers in the last stage C5 with $ 3\times 3 $ deformable convolution layers.

\begin{figure}
	\centering
	\includegraphics[scale=.60]{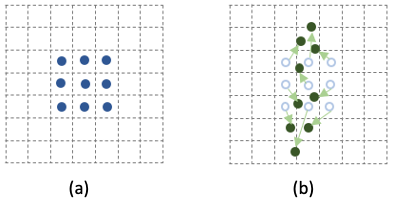}
	\caption{Comparison of $ 3\times 3 $ standard and deformable convolution sampling points. a. standard convolution sampling points (dark blue points); b. deformable convolution sampling points (dark green points).}
	\label{fig:figure5}
\end{figure}

\par Deformable convolution can adjust spatial samples with extra offsets and learn the offsets of target tasks without additional supervision. As shown in Figure 4, the two-dimensional offset can be calculated by another parallel standard convolution unit, and can also be learned end-to-end by gradient back propagation to generate new sampling positions in the feature map \cite{liu2019deformable}.  For a sampling field $R$ with the size of $ (N\times N) $,
$
R=\{(0,0),(0,1), \ldots,(N-1, N-1)\}
$, 
and an input image data $x$, for each location $ p_0 $ on the output feature map $y_{\text {deform }} $, the formula of deformable convolution is as follow \cite{liu2020visual},

\begin{equation}
y_{\text {deform }}\left(p_{0}\right)=\sum_{p_{n} \in R} w\left(p_{n}\right) * x\left(p_{0}+p_{n}+\Delta p_{n}\right)
\end{equation}

\noindent where $p_n$ is all the locations in the sampling field, and $\Delta p_{n}$ is the offset position of each sampling point, and $w(p_n )$  is the corresponding weight  \cite{liu2020visual},

\begin{equation}
\Delta p_{n}=\left(x_{offset}, y_{offset}\right) \mid (0,0), \ldots,(N-1, N-1)
\end{equation}

\noindent where $(x_{\text{offset}},y_{\text{offset}} )$ represents the offset of x coordinate and y coordinate of a certain position respectively.
\par Since the offset $\Delta p_{n}$ is typical fraction, and discrete image data cannot use non-integer coordinates, bilinear interpolation is adopted in Eq. (1). The intuitive effect is that the deformable convolution can adjust the position of sampling points according to the feature of the slender object. 
\par It is crucial to show the channel content and space location of the slender object in the image instead of the background area. The attention mechanism concentrates on the features of slender objects in the image and ignores those that are irrelevant. Hence, we propose a deformable convolution-based attention mechanism for better focusing on the features of slender objects. The proposed attention mechanism consists of two dimensions of channel and space to better extract the features of slender objects, see Figure 4. 
\par Spatial attention focuses on what spatial information is activated in the feature map. Spatial attention focuses on the spatial level of slender objects in order to enchance the valuable local spatial information while suppressing the slender objects' background noise information \cite{chen2021channel}. The spatial attention feature map is obtained by feeding the deformable C5 feature map through the spatial attention mechanism (see Figure 4). To generate two one-dimensional feature maps, the average-pooling and max-pooling are used to the deformable C5 feature map first. Second, a two-dimensional feature map is created by concatenating two one-dimensional feature maps. A novel one-dimensional spatial attention weight map is created by convolution with the $ 7\times 7 $ convolution kernel to determine the spatial attention weights of slender objects. To compress the spatial attention weights into a range, the sigmoid function is utilized $(0,1)$. Finally, the spatial attention weight map $W$ is applied to the initial feature map $y_{\text{deform}}\left(p_{0}\right)$ by element-wise multiplication to get the spatial attention feature map. The formulas are defined as follows: 

\begin{equation}
f^{\prime}=\operatorname{Avg}P_{sp}\left(y_{\text{deform}}\left(p_{0}\right)\right)\oplus\operatorname{Max}P_{sp}\left(y_{\text{deform}}\left(p_{0}\right)\right)
\end{equation}
\begin{equation}
W=\phi\left(f^{\prime}\right)
\end{equation}
\begin{equation}
F_{s p}=\sigma(W) \odot\left(y_{\text {deform }}\left(p_{0}\right)\right)
\end{equation}

\noindent where $AvgP_{\text{sp}}$ and $MaxP_{\text{sp}}$ represent average-pooling and max-pooling, respectively. The symbol $\phi$ is the convolution layer, and its filter size is $ 7\times 7 $. The symbol $\bigoplus$ is the connection operation on the channel axis. The symbol $\odot$ is the element-wise multiplication on each channel. The symbol $\sigma$ denotes the sigmoid function. 
\par Channel attention is then used to transmit the spatial attention feature map. The channel attention module sets weights to different dimensions of features so that the ones that contribute the most to the representation of slender object features is highlighted. As illustrated in Figure 4, the spatial attention feature map is first processed using the average-pooling and max-pooling layers, which can learn statistical information about the input features. Second, the pooling layer's output is processed by a shared network composed of a one-dimensional convolution layer and a fully connected layer, which is then connected by element-by-element addition. Finally, the sigmoid activation function is used. To acquire the final DFAM feature map, the learned one-dimensional channel attention weight map is applied to the spatial attention feature map via element-wise multiplication \cite{chen2021channel}. The formulas are defined as follows: 

\begin{equation}
\left.W=\sigma\left(\varphi_{1}\left(A v g P_{c h}\left(F_{s p}\right)\right)+\varphi_{2}\left(\operatorname{Max} P_{c h}\left(F_{s p}\right)\right)\right)\right)
\end{equation}
\begin{equation}
F_{c h}=W \odot F_{s p}
\end{equation}

\noindent where $AvgP_{\text{ch}}$ indicates the operation of average-pooling, while $MaxP_{\text{ch}}$ represents max-pooling.  $\varphi_1$ and $\varphi_2$ denote fully connection layers.

\subsection{Transformer encoder and decoder} 
\par Unlike convolution, which can only obtain local features, transformer employs attention-based mechanisms of encoder and decoder to obtain global to specific features \cite{carion2020end,vaswani2017attention,zhu2021ddteod,zhang2021image,han2020survey,yang2021transformer}. We make use of the transformer encoder and decoder. As input, the encoder uses multi-scale slender object feature maps. The transformer layer of encoder conducts multi-head attention to capture the context of global slender object features, which locate the association between different pixels in slender feature map. Object query is introduced in the decoder to narrow down the searching space of objects. Finally, transformer can focus on slender objects in an image. The details of transformer in object detection can be found in Deformable DETR \cite{zhu2021ddteod}. 

\subsection{Loss Function}
\par In this study, the loss function is consistent with Deformable DETR \cite{zhu2021ddteod}, and the total loss includes classification loss and regression loss. During the model training phase, the Hungarian algorithm \cite{kuhn1955hungarian} is used to match the GT with the model's prediction outcomes. The Hungarian algorithm (bipartite graph matching method) is adopted to determine the optimum arrangement with the least amount of matching loss. The optimal matching result is used to determine the loss function.

\section{Experiments and results}
\subsection{Dataset}
\par The slender objects dataset used in this study is manually extracted from MS-COCO2017 \cite{lin2014microsoft}. It includes slender objects such as toothbrush, snowboard, surfboard, etc. Totally 25,424 training images are used for training, and 1077 images for validation. Data augmentation \cite{zoph2020learning} is performed on the data through random cutting.  

\subsection{Experiments}
\subsubsection{Experimental Environment}
\par The experiment was carried out on a Xeon 3104 and an NVIDIA Tesla V100 16GB graphic card, and the environment used Pytorch 1.5.1. The network was trained for 50 epochs using the Adam optimizer \cite{da2014method}. The transformer's initial learning rate was $1\times10^{-4}$, while the backbone was $2\times10^{-5}$. The learning rate dropped by 10 times for every 20 training epochs as the number of training epochs increases. The batch size was set to 2, while the weight decay and momentum was set to 0.0001 and 0.9, respectively.

\subsubsection{Experimental Results}
\par In this study, ResNet50 was used as the backbone network. First, the pre-trained Faster R-CNN, RepPoints, FCOS, and DETR detectors were evaluated for slender object detection, see Table 1. Comparing to Faster R-CNN, RepPoint, and FCOS, it revealed that utilizing a detector based on transformer resulted in a significant improvement in the detection accuracy of slender objects. DETR boosted the $AP$ by 10.9 percent comparing to Faster R-CNN, and 14.2 percent for $AP_{\text{50}}$. DETR have significant increase on detection accuracy for both $AP_{\text{S}}$, $AP_{\text{M}}$ and $AP_{\text{L}}$. The $AP_{\text{S}}$ was increased by 5.9 percent using DETR, $AP_{\text{M}}$ was increased by 13.9 percent and $AP_{\text{L}}$ was increased by 16.5 percent comparing to Faster R-CNN. The above experiments showed the effectiveness of transformer on slender object detection.

\begin{table}[t]
	\centering
	\small
	\begin{tabular}{l|c|c|c|c|c|c}
		\toprule
		Detector & $AP$ & $AP_{\text{50}}$ & $AP_{\text{75}}$ & $AP_{\text{S}}$ & $AP_{\text{M}}$ & $AP_{\text{L}}$ \\
		\midrule
		Faster R-CNN & 17.9 &	35.1 &	15.7 &	2.7 &	17.9 &	34.5 \\
RepPoints &	18.5 &	34.1 &	18.1 &	2.6 &	18.6 &	36.1 \\
FCOS	& 20.7 &	38.6 &	20.0 &	7.5 &	24.1 &	31.0 \\
DETR	& \textbf{28.8} &	\textbf{49.3} &	\textbf{28.6} &	\textbf{8.6} &	\textbf{31.8} &	\textbf{51.0} \\
		\bottomrule
	\end{tabular}
	\caption{A comparison of the detection accuracy of four detectors for slender objects.}
	\label{table1}
\end{table}

\par Second, the dataset was trained and evaluated using DETR, Deformable DETR, and DFAM-DETR for comparing the accuracy of slender object detection. ResNet50 was employed for the backbone network. We initialize our backbone networks with the weights pre-trained on ImageNet \cite{deng2009large,krizhevsky2012imagenet}. Transformer is trained with random initialization. Results showed improvement in detection accuracy on $AP$, $AP_{\text{50}}$, and $AP_{\text{75}}$ with DFAM-DETR, see Table 2. Comparing to DETR, the proposed DFAM-DETR increases slender objects detection accuracy by 4.6 percent on $AP$, and 4.3 percent on $AP_{\text{50}}$. DFAM-DETR outperformed Deformable DETR by 2 percent increase on $AP$ and 2.2 percent increase on $AP_{\text{50}}$. Furthermore, DFAM-DETR significantly improves detection accuracy for small and medium objects. The results revealed that accuracy was improved by 4.1 and 2.6 percent for small objects, while 6.3 and 2.8 percent increase for medium objects. However, the detection accuracy for large objects was dropped by 0.6 percent comparing to Deformable DETR. Accuracy in detecting larger objects may reduce as a result of insufficient dimensionality exploration, receptive field susceptibility limitations, or the absence of the convolution kernel for larger receptive fields.

\begin{table}[t]
	\centering
	\small
	\begin{tabular}{l|c|c|c|c|c|c}
		\toprule
		Detector & $AP$ & $AP_{\text{50}}$ & $AP_{\text{75}}$ & $AP_{\text{S}}$ & $AP_{\text{M}}$ & $AP_{\text{L}}$ \\
		\midrule
		DETR &	30.8 &	52.7 &	30.3 &	11.5 &	33.7 &	53.0 \\
		Deformable DETR &	33.4 &	54.8 &	34.3 &	13.0 &	37.2 &	\textbf{54.5} \\
		DFAM-DETR &	\textbf{35.4} &	\textbf{57.0} &	\textbf{37.8} &	\textbf{15.6} &	\textbf{40.0} &	53.9 \\
		\bottomrule
	\end{tabular}
	\caption{A comparison of the detection accuracy of detectors that based on transfomer.}
	\label{table1}
\end{table}

\par The detection accuracy and convergence curves of Deformable DETR and proposed DFAM-DETR are illustrated in Figure 6. DFAM-DETR shows higher slender detection accuracy of $AP$ comparing to Deformable DETR. Moreover, DFAM-DETR achieves 2 times less training epochs. As shown in Figure 7, the training loss of DFAM-DETR is significantly lower than Deformable DETR. Again, it depicts that DFAM-DETR has superior convergence speed and it has better performance on slender object detection.

\begin{figure}
	\centering
	\includegraphics[scale=.60]{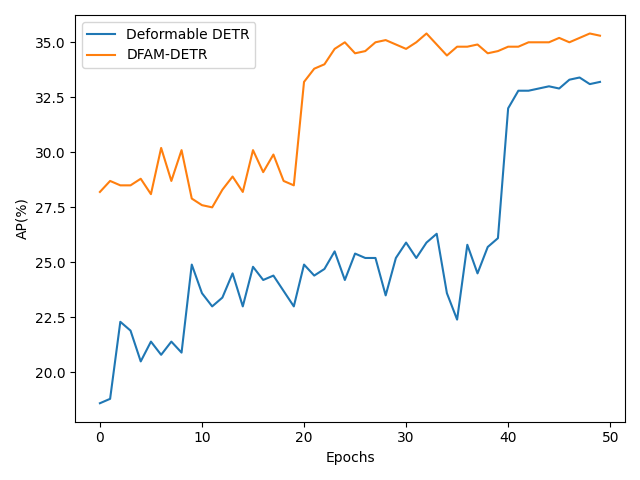}
	\caption{A comparison of accuracy and epochs between Deformable DETR and DFAM-DETR}
	\label{fig:figure6}
\end{figure}

\begin{figure}
	\centering
	\includegraphics[scale=.60]{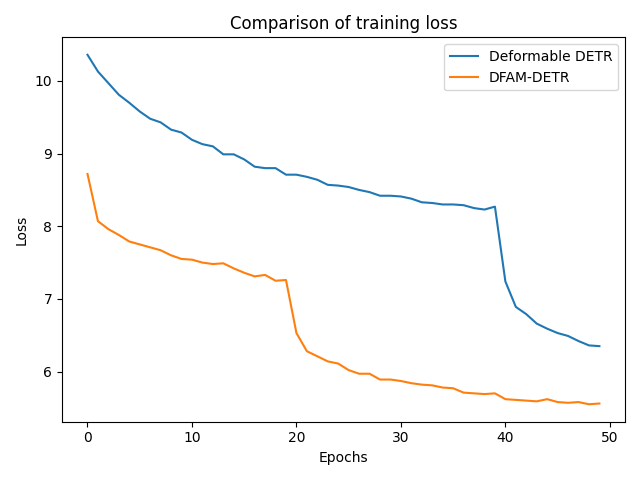}
	\caption{A comparison of loss and epochs between Deformable DETR and DFAM-DETR}
	\label{fig:figure7}
\end{figure}

\section{Conclusion}
\par This study proposes DFAM-DETR, a slender object detector based on Deformable DETR. Comparing to other popular detectors, it delivers greater detection accuracy for slender objects. With the proposed DFAM's deformable convolution and attention mechanism, it overcomes the limitation of convolution with fixed sampling points for slender object detection. DFAM-DETR detector improves the feature extraction ability of slender objects with greater detection accuracy and convergence speed. The research of DFAM-DETR will be expanded to include its detection capability and performance on various size of slender objects.

\newpage
\bibliographystyle{named}
\bibliography{DFAM_DETR}

\end{document}